# A system for exploring big data: an iterative k-means searchlight for outlier detection on open health data

A. Ravishankar Rao, PhD,
Fellow, IEEE
Fairleigh Dickinson University, NJ, USA
raviraodr@gmail.com

Daniel Clarke, M.S.,
Student Member IEEE
Fairleigh Dickinson University, NJ, USA
danieljbclarke@gmail.com

Subrata Garai
IT Software Engineer
subratagarai@gmail.com

Soumyabrata Dey, PhD.
soumyabrata.dey@gmail.com

*Abstract*

**The interactive exploration of large and evolving datasets is challenging as relationships between underlying variables may not be fully understood. There may be hidden trends and patterns in the data that are worthy of further exploration and analysis.**

**We present a system that methodically explores multiple combinations of variables using a searchlight technique and identifies outliers. An iterative k-means clustering algorithm is applied to features derived through a split-apply-combine paradigm used in the database literature. Outliers are identified as singleton or small clusters. This algorithm is swept across the dataset in a searchlight manner. The dimensions that contain outliers are combined in pairs with other dimensions using a susbset scan technique to gain further insight into the outliers.**

**We illustrate this system by anaylzing open health care data released by New York State. We apply our iterative k-means searchlight followed by subset scanning. Several anomalous trends in the data are identified, including cost overruns at specific hospitals, and increases in diagnoses such as suicides. These constitute novel findings in the literature, and are of potential use to regulatory agencies, policy makers and concerned citizens.**

## I. INTRODUCTION AND MOTIVATION

The intersection of the areas of big data and health care provide many opportunities for innovation. One of the interesting directions concerns open health data initiatives, which have been underway at several national and regional agencies worldwide, including the NHS in the UK, the Center for Medicare and Medicaid Services (CMS) in the USA, and New York State Statewide Planning and Research Cooperative System (SPARCS) [1-3]. This effort is driven by transparency as citizens are demanding more accountability from their governments.

Rao et al pointed out many challenges that need to be overcome [4-8], including the cleaning of these datasets, the difficulty in combining information from multiple agencies and sources, and the lack of a single platform to perform end-to-end data analytics. Though open data initiatives are expected to help the citizenry, there is a large gap between the ability of citizens to use existing tools and the sophisticated types of analyses that need to be performed. For instance, without aggregating information from multiple years, it is not easy to determine cost trends for different procedures, or to be able to identify individual hospitals that may be good for treatment of a specific condition.

Given the proliferation of different news sources and articles for information [9] it is becoming increasingly difficult to verify the "ground truth" in several domains. For instance, in the area of health care, we may want to know the expected cost of a given procedure in a given geographic area, such as hip replacement in New York city. Such cost information is difficult to obtain directly from the providers in countries like the US. However, open health data provides a source of relatively unbiased information [1-3]. This enables researchers and citizens to probe the the health care system and gain an unbiased view of costs.

When presented with a vast amount of data, important questions to consider concern the regularity of the data, and conversely any types of irregularities [10]. Both these questions lead to interesting insights, and are required for model building and machine learning [11].

In a companion paper submitted to IJCNN 2018 [12], we investigate a model building exercise for the prediction of costs based on open health data. The focus of the current paper is on irregularities or outliers. Both types of analyses are important to gain a complete perspective about the data. The detection of outliers is particularly challenging as there is no standardized definition of what an outlier is [13].

The building of models for health data is challenging as the distribution of many variables such as cost in health care is heavily tailed [14], and different diseases may be associated with their own unique distribution types, e.g. Weibull, Cox or Tobit. Consequently, some researchers have thresholded the cost data, and ignore costs above say $50,000 in order to fit an appropriate model such as linear regression to the data [15]. If we simply apply such a technique and threshold the data to produce outliers, we could generate far too many possibilities that can be screened carefully.

There are many reasons that make outlier detection an important problem. Firstly, there may be trending health conditions that are not receiving sufficient attention from health providers or parents or educators. An example of this is the high increase in mental health issues with the teenage population in New York state [7]. Secondly, there may be specific hospitals that are underperforming relative to their peers. Identification of anomalous cost patterns could alert regulatory authorities, and also potential patients who may be considering treatment at these hospitals. Thirdly, outliers may be an indication of outright fraud such as gross overcharging for certain medical procedures. Fraud identification is an important application area for outlier detection [16].

We caution that the detection of an outlier does not directly imply that it represent fraudulent activity. Rather, it is merely a candidate for further consideration. Leite [17] presented a visual analytics approach to identify outliers in the financial domain, where candidates are presented to a human to enable fine tuning and generation of automatic alarms. Our approach is similar, where we utilize an outlier detection algorithm embedded in a human-in-the-loop system for verification.

In our earlier work, [8] we combined techniques from database analysis and machine learning to derive an iterative k-means technique to identify outliers. We developed a proof of concept for the technique, where it was applied successfully to understand graduation trends in different healthcare specialties using data from CMS [18] for 800,000 practitioners. The main contribution of the current paper is to enhance the iterative k-means algorithm by using searchlight and subset scanning techniques, and to apply the enhanced algorithm on much larger datasets, containing de-identified data from approximately 15 million patients a year over a 5 year period. This data is provided by the SPARCS program, and contains discharge information about disease diagnoses and costs [3]. Our technique is described in detail in Section IV.

## II. BACKGROUND AND RELATED WORK

Gregory et al [14] reviewed several models for healthcare costs, including Weibull, Cox, Aalen and Tobit. They determined that it is not possible for a single model to deal with the variety that is encountered in cost data, and consequently developed separate models for different disease conditions. Mihaylova [19] presents an overview of different statistical methods to analyze data in the area of healthcare, and reaches a similar conclusion to Gregory et al [14] in that different models should be applied in different scenarios. Hence there is a need to investigate an approach that does not assume a prior statistical distribution for the cost values. The approach we present in the current paper is model-free.

Chen et al. [20] advocate a visualization process to scan data and understand relationships between the existing variables. A purely visual exploration approach is limited, as many combinations of variables need to be explored manually. We facilitate exploration by applying a clustering algorithm to automatically identify outliers and highlight interesting relationships.

Goldstein and Uchida [21] review techniques for unsupervised anomaly detection. Gupta et al. [22] present an overview of outlier detection techniques. Hauskrecht [23] presents a system developed at the University of Pittsburgh for monitoring individual patient data and issuing alerts about unusual case decisions. For this, nerarly 4500 electronic health records (EHRs) of patients were utilized. In contrast, we use a significantly larger dataset covering several million patient records, but at a coarser scale than their EHRs. The research presented in the current paper shows that one is able to extract meaningful outlier information from large public healthcare datasets. Krumholz [24] has pointed out that analysis is the bottleneck in the learning process in health-care. Hence, there is a need for a technique to quickly detect meaningful trends and outliers, and to present them to decision makers for further action. Our current paper presents a technique to address this need.

## III. DESIGN

In our previous work, we investigated the architecture of an open-source system for the analysis of open health data[5, 6]. We used a Python-based solution consisting of the following components: Python Pandas, Scikit-Learn and Matplotlib [5]. The Scikit-Learn Python library [25] provides several machine-learning capabilities such as clustering, classification and prediction.

In previous work [8], we presented an iterative k-means algorithm for outlier detection. We extend this work with the following additional enhancements, as shown in Figure 1. Firstly, we apply the iterative k-means algorithm in a searchlight manner, where the outlier detector is swept across pairwise aggregations of variables. This is similar to the use of searchlights in the functional magnetic resonance imaging literature in order to identify interesting patterns of brain activity, as explored by Rao [26]. Secondly, we apply a subset scan technique to further elaborate any dimensions that contain outliers. This allows more detailed relationships between the variables to be teased out of the data. In this paper we apply this algorithm to the SPARCS dataset and examine the results generated by using different aggregation possibilities, denoted by "Aggregate 1", "Aggregate 2" and so on, shown in **Figure 1**.

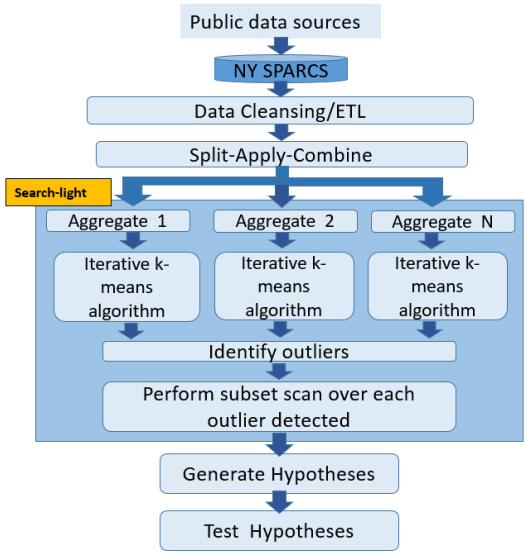

Figure 1: Proposed pipeline for data analysis involving outlier detection using a searchlight, followed by a subset scan and post-processing.

The availability of a fast and robust outlier detection technique like the iterative k-means algorithm enables several interesting use cases, two of which are shown in Figure 2. Regulators can use the outliers as hypotheses for further analysis that needs to be investigated. Policy makers can use spikes in occurrences of certain disease patterns to allocate resources to treat them. Individuals can use the information on outliers as cautionary flags to consider when considering their medical treatment options. They could use the identification of outliers as a basis to conduct further investigation, say by searching news sources for further information about the outliers.

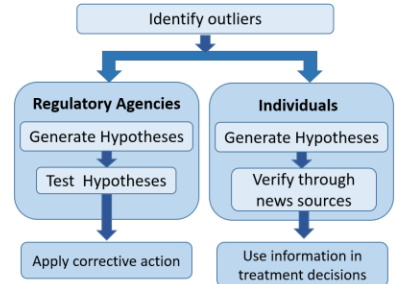

**Figure 2:** Use case scenarios enabled by the outlier detection algorithm.

We use the SPARCS dataset, consisting of de-identified in-patient discharge information about disease diagnoses and costs [3]. The aggregate data [2] contains 15,213,123 rows of patient data from 254 hospitals and 58 counties. There are a total of 264 different diagnosis descriptions.

In order to perform the aggregation in Figure **1** we can use different features in the original dataset, such as hospital county, CCS Diagnosis Description, or Facility Name, which are illustrated in Figure 3.

| Hospital County | Albany |
|---|---|
| Facility Name | Albany Med. Center |
| Age Group | 18 to 29 |
| Zip Code - 3 digits | 124 |
| Gender | F |
| Race | Other Race |
| Ethnicity | Unknown |
| CCS Diagnosis Description | OTH COMP BIRTH |
| CCS Procedure Description | CESAREAN SECTION |
| Payment Typology 1 | Managed Care |
| Total Costs | $12,068.11 |

Figure 3: An example of the data fields in SPARCS, spanning multiple features. A few selected fields are shown, drawn from a total of 35 such fields.

Sample Diagnosis Descriptions consist of strings such as ABDOMINAL HERNIA, ABDOMINAL PAIN, ACQUIRD FOOT DEFORMIT and ACUTE CVD.

## IV. METHODS

We review the **iterative k-means** algorithm from Rao and Clarke [8] as shown in Figure 4. The algorithm repeatedly runs the k-means clustering technique. During each run, smaller clusters of points are treated as outliers and removed. This process is repeated until no small clusters are present, or until a fixed number of iterations are executed. The resulting clusters capture the most relevant groups after outlier removal. Figure 4 illustrates this procedure graphically through hypothetical data and shows four iterations labeled Step (a) - (d). We first select the number of clusters, e.g. k=4 and proceed as below.

1. Apply the k-means clustering algorithm.
2. If single-element or substantially small clusters (e.g. size < 2 or 3) exist, treat these as outliers and remove them from the dataset. Continue the computation with the remaining data.
3. If no substantially small clusters remain, the iterative k-means algorithm is terminated.

This approach exploits the sensitivity of the k-means algorithm towards outliers in order to isolate them. Our approach requires the number of clusters to be specified, but there is no limit on the number of outliers that can exist at each iteration. Our algorithm is executed for multiple iterations until the user is satisfied with the results. Sample code is presented in Rao [8].

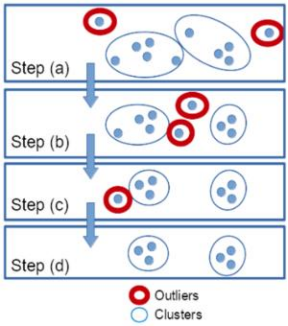

**Figure 4:** The iterative k-means algorithm works sequentially on the original data, shown in steps (a)-(d). Individual data points are shown as blue dots. The red circles are outliers that are not considered for further analysis. The blue ellipses identifiy the remaining clusters that are then processed in the next iteration.

The processing pipeline in Figure **1** consists of the following steps:

1. Use the split-apply-combine paradigm [27]. We use the Pandas *pivot table* functionality to group the data by different feature columns in the database, as follows.
2. This produces grouped items' counts for each "*CCS Diagnosis Description*", which are then further binned over the field *"Discharge Year"*. The result is a feature vector for each specialty which consists of the count of hospitalization incidents binned into 'Year of the incident'. Then this feature vector is scaled to find the percentage of change with respect to base year (2009) values. Below is the graphical representation of the matrix created.

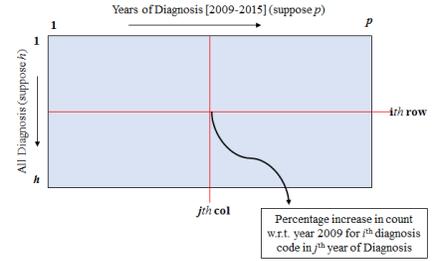

3. The feature vectors for each specialty are then processed by the iterative k-means algorithm outlined in Figure 4.
4. We use interactive visualization to present different graphs to the user, which permits an understanding of unusual trends in the data. This approach is illustrated in Figure 5.

Aggregation with respect to a single field can hide interesting data patterns which may be visible when analyzed with respect to the other fields. At the other extreme, aggregation with respect to all possible fields subdivides the data into the highest level of granularity, and significantly increases the search space. In search of a balance we propose a novel algorithm combining a subset scan [28] with the iterative k-means outlier algorithm. If an outlier is found for the aggregation over a given field, say *N*, then we examine subsets that include the field *N*. For instance, if we find an outlier in counts for Diagnoses codes for "Suicides", we then search for outliers in "Suicides" over "Age Groups", "Ethnicity", "County" and so on. For each of these subsets, the iterative K-means algorithm is used. In summary, we starting with the entire data set, and hierarchically focus our search on only the outlier points to investigate the actual subset of fields responsible for the unusual patterns.

The current paper presents significant improvements over [8] in the size of the dataset considered (more than 15 million entries), a larger number of descriptive features, different choices for aggregation of the data, and a subset scan procedure to investigate outliers in more detail.

V. RESULTS

We applied the iterative k-means algorithm described in Section IV with k=8 on the SPARCS dataset. We explored different aggregations including total costs by hospital, or by county. Similarly, we aggregated total costs according to the individual CCS Diagnosis Descriptions. Trends in cost increases can also be computed. The results are organized by the types of aggregation applied, a few of which are shown below.

## A. Percentage increase in counts of incidences per year, aggregated over CCS Diagnosis Descriptions

We computed the percentage change in total counts of incidences for medical procedures reported to New York State from the years 2009-2014 by using the year 2009 as the baseline. The results are shown in the order of sequential outlier detection, followed by plots of the remaining clusters. The significance of these results is reviewed in the discussion section.

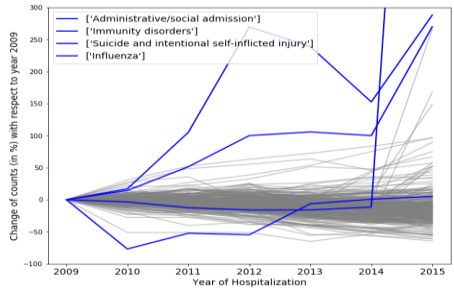

Figure 5: Each curve in this figure represents the percentage change in count for treatment of a specific CCS Diagnosis Description relative to the year 2009. The outliers detected at each iteration are shown in blue.

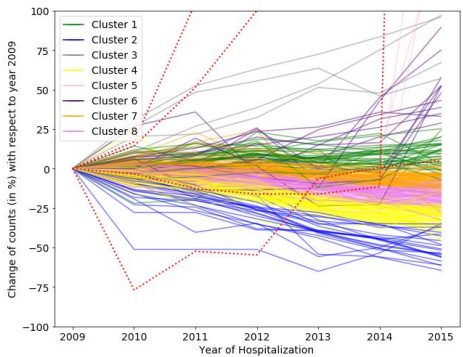

**Figure 6:** This figure shows the clusters in the count trends for different CCS Diagnosis Descriptions. The top 4 outliers are shown in dotted red lines, and the k-means algorithm is run on the remaining trend curves, producing the color-coded clusters.

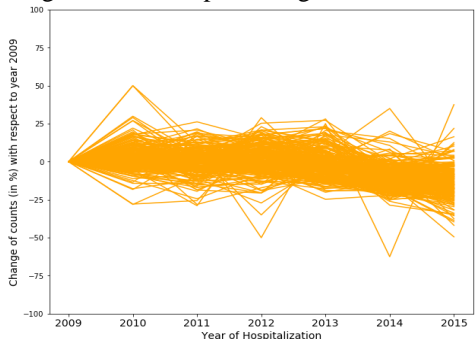

**Figure 7:** This figure shows the largest cluster, with 66 members, when running the k-means algorithm after the first five outliers are removed.

## B. Subset Scan

Once interesting outliers are identified, we perform a subset scan where we further refine the dimensions over which aggregation is performed. We use the same 'iterative k-means' method to include other dimensions of the dataset, one at a time, along with the current feature space. The following 2D matrix is created, where the 'Age Group' is considered as the second dimension.

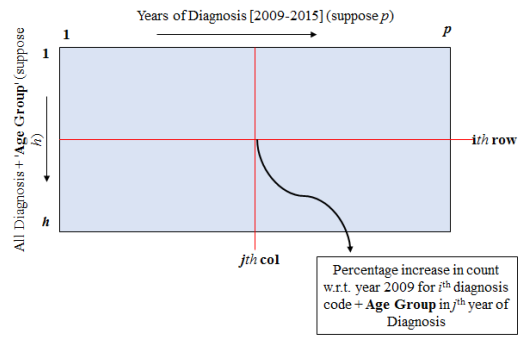

We note that if we had N distinct values of Diagnosis Codes and T number of years, the input used in the first experiment would contain a [N x T] matrix. Whereas using another dimension (e.g. – 'Age Group', 'Race', 'Ethnicity' etc.) along with the feature space would produce a [N x T x S] matrix, considering the third dimension has S unique values. Hence, we created a list of the other dimensions which might contribute to the outliers already found and developed count matrices for each aggregation in the list. The results obtained are shown in Figure **8**.

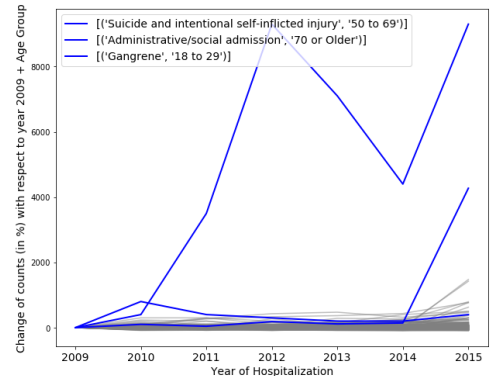

(a) Subsetting the aggregate counts by Age group.

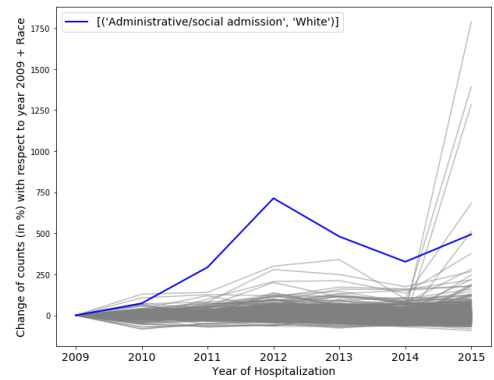

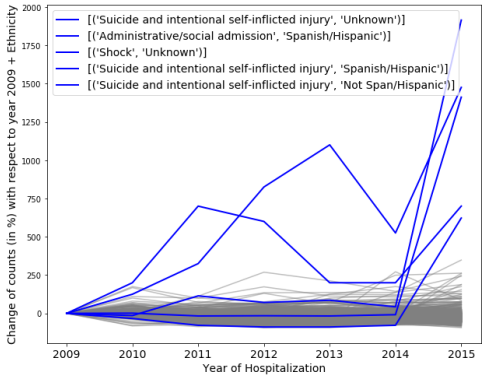

(b) Subsetting the aggregate counts by Race.

(c) Subsetting the aggregate counts by Ethnicity.

Figure 8: This illustrates the subset scan procedure where we explore additional dimensions related to existing outliers. Examples are shown for subsetting by Age group, Race and Ethnicity, which are the subsets that produce outliers.

## C. Total costs aggregated by CCS Diagnosis Description over all the Hospitals

Figure 9 shows the results of aggregating costs by CCS Diagnosis Description over all the hospitals, followed by the iterative k-means algorithm. The results are shown in the order of sequential outlier detection, followed by plots of the remaining clusters.

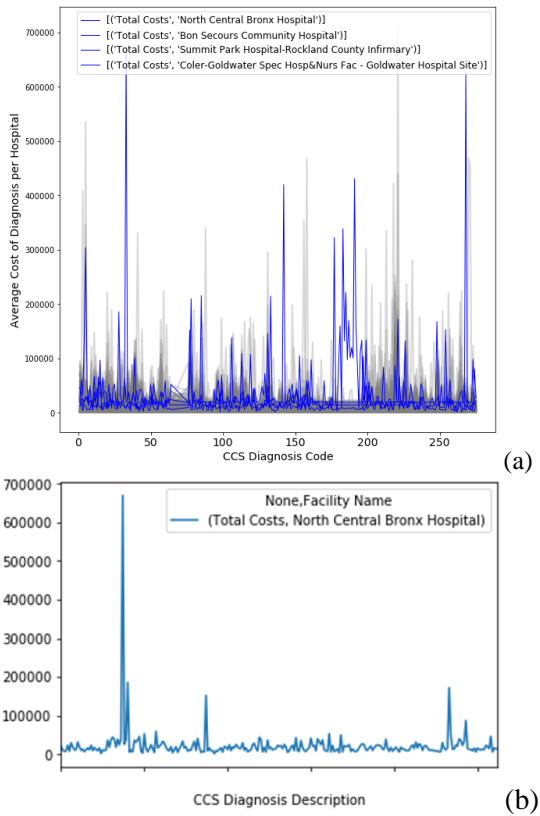

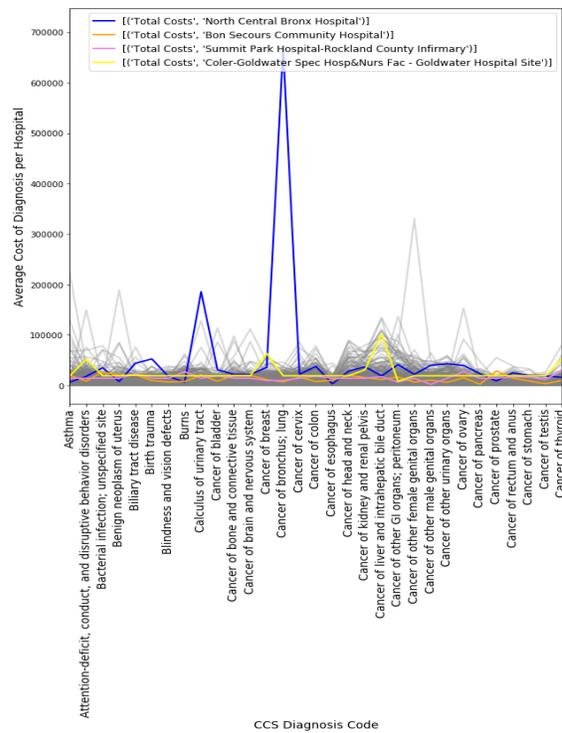

**Figure 9:** The x-axis consists of the index number assigned to each CCS Diagnosis Description. Each continuous curve represents a single hospital. (a) shows outliers in the distribution of "Total Costs", which identifies hospitals such as the Bon Secours Community Hospital, North Central Bronx Hospital and Henry Carter Specialty Hospital. (b) shows the pattern of cost for an individual hospital, the North Central Bronx Hospital. (c) shows a zoomed-in version of the costs for North Central Bronx Hospital, showing a spike for lung cancer.

From Figure 9 we can see that outliers consist of combinations of spikes in various dimensions. Though we could identify spikes in one dimension at a time, the advantage of the iterative k-means clustering algorithm is that it looks for outliers in the pattern of the entire cost distribution across all CCS Diagnosis Descriptions. We used 6 iterations of the iterative k-means algorithm to identify these outliers.

We have applied our algorithm to detect outliers in several additional combinations of variables, including costs across counties, which identified Westchester county as having the largest cost increase for mental health diseases [7]. These results cannot be shown due to space limitations.

## VI. DISCUSSION

From Figure 5 we observe that the following Diagnosis descriptions are outliers: 'Administrative/social admission', 'Immunity disorders', 'Suicide and intentional self-inflicted injury' and 'Influenza'. Specifically, 'Suicide and intentional self-inflicted injury' has a sharp spike for year 2015 compared to other years.

Figure 8 is produced by adding three new dimensions ('Age Group', 'Race', 'Ethnicity') to the outlier detection method. This provides a more nuanced analysis of the trends in the data. Specifically, we note that 'Suicide and intentional self-inflicted injury' was high in Age Group '50 to 69' in the year 2015. The analysis also shows that this same diagnosis is high in the year 2015 for 'Spanish/Hispanic' and 'Non Spanish/Hispanic' ethnicities. Furthermore, there is an anomaly for 'Administrative/social admission' when viewed for the 'white' race.

Figure 9 identifies four hospitals where the average costs of different diagnoses are anomalous with respect to the rest of the hospitals. By zooming into the diagnoses descriptions, we can see that "North Central Bronx Hospital" has very high average cost for diagnosis 'Cancer of Bronchus; Lung'. Similarly, 'Summit Park Hospital' has very high average cost for 'personality disorder' treatment. Also, 'Bon Secours Community Hospital' has high treatment cost for a several diagnoses. This suggests that a targeted intervention to educate the public in the North Bronx area about lung cancer issues may be useful in preventing incidences of this disease.

It is interesting to observe that a Big Data approach with interactive visualization tools can provide researchers with a ready capacity to start with the raw data provided by the government agencies and quickly identify and explore interesting trends. We expect the widespread adoption of such tools to enable concerned citizens to draw their own conclusions from important national data sources without having to deal with potential biases in reporting.

The Summit Park Hospital identified earlier was closed in 2015 due to cost overruns [29]. A recent news article mentions a hospital consolidation between Westchester Medical Center and the Bon Secours Health System [30]. It is likely that the outlier cost distribution in Figure 9 was an indication of issues at the Bon Secours Health System, and this hospital became a candidate for merger talks. This is merely a hypothesis at this point, driven directly by the data. Further verification is required. Nevertheless, the data provides a good source of possible hypotheses about the performance of different hospitals, which can then be subject to further scrutiny. We are using these examples to illustrate the verification process outlined in Figure 2.

Figure 9 sheds light on the developments surrounding the creation of the new Henry J. Carter Specialty Hospital in 2011 [31]. This hospital was formed to provide long-term acute care for physically disabled and medically fragile individuals. Given the type of patients that this hospital is designed to serve, it appears reasonable that their cost structure will be different from other hospitals in New York state.

These results demonstrate that we can quickly determine interesting and relevant trends in large health-care related datasets. This capability could provide concerned citizens with an unbiased data-driven interpretation of breaking news events in their regions as well as nationally.

In order to facilitate widespread adoption of the techniques presented in this paper, we have made our framework and code available freely to the research community at github.com/fdudatamining/.

## VII. CONCLUSION

We presented an open-source toolkit based on Python that can be utilized to analyze and interpret large datasets thereby driving insight. Many government agencies, such as Medicare in the US release detailed data about their inner workings, but the capabilities of tools to interpret this data has not kept pace. Since the relationships between variables in these large datasets are not fully known, users typically engage in visual exploration, which tends to be slow and manually intensive. We have developed a machine learning approach, called iterative k-means, where clusters and outliers are automatically identified and presented to the user. This facilitates rapid visual exploration of new datasets.

We applied our toolkit to analyze health care data released by New York State SPARCS. Our technique identified interesting and meaningful trends in counts and cost increases for different diagnoses such as lung cancer and suicide rates. This information can be utilized by policy makers for targeted interventions to improve public health in specific regions. Our approach should also prove valuable to other researchers, and concerned citizens who are interested in exploring open health data.

REFERENCES:

[1] "http://www.medicare.gov/hospitalcompare/data/total-performance-scores.html."
[2] "https://health.data.ny.gov/Health/Hospital-Inpatient-Discharges-SPARCS-De-Identified/rmwa-zns4."
[3] *New York State Department Of Health, Statewide Planning and Research Cooperative System (SPARCS)*. Available: https://www.health.ny.gov/statistics/sparcs/

[4] A. R. Rao and D. Clarke, "Facilitating the Exploration of Open Health-Care Data Through BOAT: A Big Data Open Source Analytics Tool," in *Emerging Challenges in Business, Optimization, Technology, and Industry*, ed: Springer, 2018, pp. 93-115.

[5] A. R. Rao and D. Clarke, "A fully integrated open-source toolkit for mining healthcare big-data: architecture and applications," in *IEEE International Conference on Healthcare Informatics ICHI*, Chicago, 2016, pp. 255-261.

[6] A. R. Rao, A. Chhabra, R. Das, and V. Ruhil, "A framework for analyzing publicly available healthcare data," in *2015 17th International Conference on E-health Networking, Application & Services (IEEE HealthCom)*, 2015, pp. 653-656.

[7] A. R. Rao and D. Clarke, "Hiding in plain sight: insights about health-care trends gained through open health data," *Journal of Technology in Human Services,* January 19, 2018 2018.

[8] A. R. Rao and D. Clarke, "An open-source framework for the interactive exploration of Big Data: applications in understanding health care " presented at the IJCNN, International Joint Conference on Neural Networks, 2017.

[9] W. L. Bennett, *News: The politics of illusion*: University of Chicago Press, 2016.

[10] S. Kirby, "Spontaneous evolution of linguistic structure-an iterated learning model of the emergence of regularity and irregularity," *IEEE Transactions on Evolutionary Computation,* vol. 5, pp. 102-110, 2001.

[11] I. H. Witten, E. Frank, M. A. Hall, and C. J. Pal, *Data Mining: Practical machine learning tools and techniques*: Morgan Kaufmann, 2016.

[12] A. R. Rao and D. Clarke, "A comparison of models to predict medical procedure costs from open public healthcare data," 2018.

[13] C. C. Aggarwal, "Outlier analysis," in *Data mining*, 2015, pp. 237-263.

[14] D. Gregori, M. Petrinco, S. Bo, A. Desideri, F. Merletti, and E. Pagano, "Regression models for analyzing costs and their determinants in health care: an introductory review," *International Journal for Quality in Health Care,* vol. 23, pp. 331-341, 2011.

[15] R. B. Cumming, D. Knutson, B. A. Cameron, and B. Derrick, "A comparative analysis of claims-based methods of health risk assessment for commercial populations," *Final report to the Society of Actuaries,* 2002.

[16] V. Hodge and J. Austin, "A survey of outlier detection methodologies," *Artificial intelligence review,* vol. 22, pp. 85-126, 2004.

[17] R. A. Leite, T. Gschwandtner, S. Miksch, E. Gstrein, and J. Kuntner, "Visual analytics for fraud detection: focusing on profile analysis," in *Proceedings of the Eurographics/IEEE VGTC Conference on Visualization: Posters*, 2016, pp. 45-47.

[18] "https://data.medicare.gov/Physician-Compare/National-Downloadable-File/s63f-csi6."

[19] B. Mihaylova, A. Briggs, A. O'hagan, and S. G. Thompson, "Review of statistical methods for analysing healthcare resources and costs," *Health economics,* vol. 20, pp. 897-916, 2011.

[20] M. Chen, D. Ebert, H. Hagen, R. S. Laramee, R. Van Liere, K.-L. Ma*, et al.*, "Data, information, and knowledge in visualization," *IEEE Computer Graphics and Applications,* vol. 29, pp. 12-19, 2009.

[21] M. Goldstein and S. Uchida, "A comparative evaluation of unsupervised anomaly detection algorithms for multivariate data," *PloS one,* vol. 11, p. e0152173, 2016.

[22] M. Gupta, J. Gao, C. C. Aggarwal, and J. Han, "Outlier detection for temporal data: A survey," *IEEE Transactions on Knowledge and Data Engineering,* vol. 26, pp. 2250-2267, 2014.

[23] M. Hauskrecht, I. Batal, M. Valko, S. Visweswaran, G. F. Cooper, and G. Clermont, "Outlier detection for patient monitoring and alerting," *Journal of Biomedical Informatics,* vol. 46, pp. 47-55, 2013.

[24] H. M. Krumholz, "Big data and new knowledge in medicine: the thinking, training, and tools needed for a learning health system," *Health Aff (Millwood),* vol. 33, pp. 1163-70, Jul 2014.

[25] F. Pedregosa, G. Varoquaux, A. Gramfort, V. Michel, B. Thirion, O. Grisel*, et al.*, "Scikit-learn: Machine learning in Python," *Journal of Machine Learning Research,* vol. 12, pp. 2825-2830, 2011.

[26] A. R. Rao, R. Garg, and G. A. Cecchi, "A spatio-temporal support vector machine searchlight for fMRI analysis," in *Biomedical Imaging: From Nano to Macro, 2011 IEEE International Symposium on*, 2011, pp. 1023-1026.

[27] H. Wickham, "The split-apply-combine strategy for data analysis," *Journal of Statistical Software,* vol. 40, pp. 1-29, 2011.

[28] D. B. Neill, "Fast subset scan for spatial pattern detection," *Journal of the Royal Statistical Society: Series B (Statistical Methodology),* vol. 74, pp. 337-360, 2012.

[29] R. Brum, "Offer rejected, Summit Park closing moves forward," in *Journal News (lohud.com)*, ed, 2015.

[30] "Downgraded: Westchester Medical Center's credit rating drops," ed. http://www.lohud.com/story/news/local/2015/05/12/westchester-medical-center-credit-rating-downgraded/27201735/, 2015.

[31] I. Michaels, "Henry J. Carter Specialty Hospital and Nursing Facility Receives First Patients As Goldwater Campus on Roosevelt Island Nears Closing," in *NYC Health+Hospitals*, ed. https://www.nychealthandhospitals.org/pressrelease, 2013.